\documentclass[10pt,twocolumn,letterpaper]{article}

\usepackage{cvpr}              % To produce the CAMERA-READY version

\usepackage{booktabs}
\usepackage{tcolorbox}
\tcbuselibrary{listingsutf8}
\usepackage{tcolorbox}
\usepackage{graphicx} % Optional, for figure environment

\definecolor{cvprblue}{rgb}{0.21,0.49,0.74}
\usepackage[pagebackref,breaklinks,colorlinks,allcolors=cvprblue]{hyperref}
\usepackage[capitalize]{cleveref} 

\def\paperID{29} % *** Enter the Paper ID here
\def\confName{CVPR}
\def\confYear{2025}

\title{DPO Learning with LLMs-Judge Signal for Computer Use Agents}

\author{
Man Luo\textsuperscript{1} \quad
David Cobbley\textsuperscript{1} \quad
Xin Su\textsuperscript{1} \quad
Shachar Rosenman\textsuperscript{1} \quad
Vasudev Lal\textsuperscript{1} \quad \\
Shao-Yen Tseng\textsuperscript{1} \quad
Phillip Howard\textsuperscript{2} \\
\textsuperscript{1}Intel \quad
\textsuperscript{2}Thoughtworks \\
{\tt\small
\{xin.su, man.luo, david.j.cobbley, shachar.rosenman, vasudev.lal,shao-yen.tseng\}@intel.com} \\
{\tt\small phillip.howard@thoughtworks.com}
}

\begin{document}
\maketitle
\begin{abstract}
Computer use agents (CUA) are systems that automatically interact with graphical user interfaces (GUIs) to complete tasks. CUA have made significant progress with the advent of large vision-language models (VLMs). However, these agents typically rely on cloud-based inference with substantial compute demands, raising critical privacy and scalability concerns, especially when operating on personal devices. In this work, we take a step toward privacy-preserving and resource-efficient agents by developing a lightweight vision-language model that runs entirely on local machines. To train this compact agent, we introduce an LLM-as-Judge framework that automatically evaluates and filters synthetic interaction trajectories, producing high-quality data for reinforcement learning without human annotation. Experiments on the OS-World benchmark demonstrate that our fine-tuned local model outperforms existing baselines, highlighting a promising path toward private, efficient, and generalizable GUI agents.
\end{abstract}
    
\section{Introduction}
\label{sec:intro}

Computer Use Agents (CUAs) aim to automate user interactions with graphical user interfaces (GUIs), offering potential to revolutionize how humans interact with software~\cite{deng2023mind2web, zhou2023webarena, xie2024osworld, agashe2025agent}. These agents can perform a wide range of tasks such as navigating applications, clicking buttons, entering data, and monitoring workflows, making them valuable tools for software testing, digital productivity, accessibility, and even robotic process automation. The emergence of large vision-language models (VLMs)~\cite{QwenVL2023, liu2023visual} has recently accelerated progress in this space, as VLMs can jointly reason over visual elements and natural language instructions, making them ideal candidates for GUI understanding and interaction.

Despite this progress, current state-of-the-art GUI agents typically rely on cloud-hosted VLMs with large computational workload. These systems require GPU-backed infrastructure, fast network connections, and remote data transfer, raising significant concerns about latency, cost, scalability, and most critically, privacy, especially when interacting with personal data on user-facing applications. 

To address these limitations, we present a privacy-preserving and resource-efficient GUI agent designed to operate entirely on local machines. Our agent is built on a compact vision-language model that maintains task competence while fitting within the constraints of consumer hardware. This local-first design eliminates the need for network communication during inference, providing strong privacy guarantees and enabling deployment in security-critical or bandwidth-limited environments.

\begin{figure*}[htbp]
\centering
\includegraphics[width=0.9\textwidth]{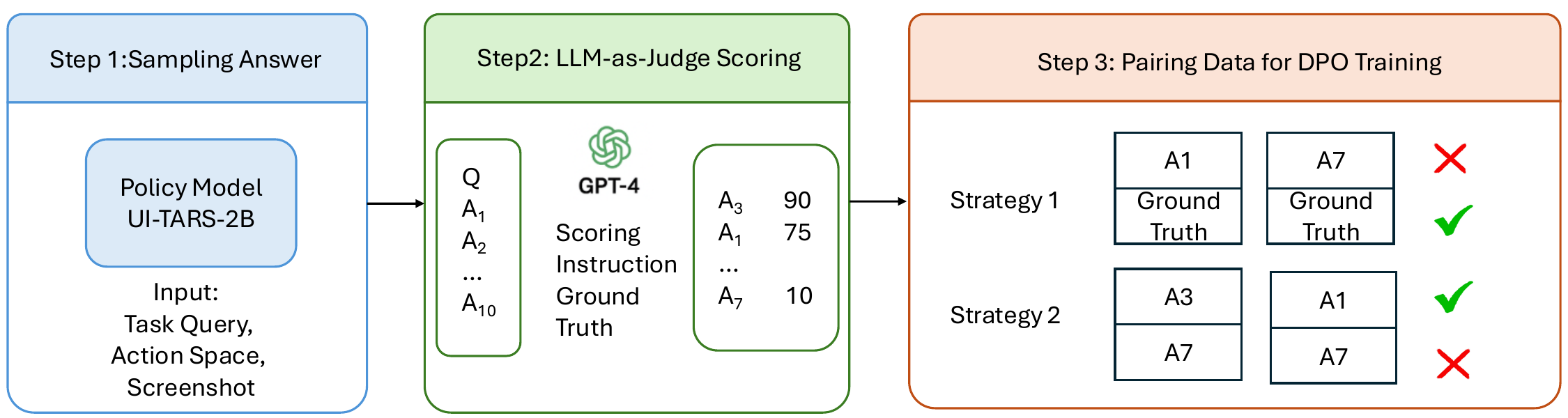}
\caption{Pipeline of creating DPO training data for the policy model (UI-TARs-2B) using LLM-as-judge method.}
\label{fig:cua}
\end{figure*}

Training such a lightweight agent, however, poses a major challenge: how to obtain high-quality GUI interaction data without extensive human labeling. To solve this, we propose an LLM-as-Judge framework~\cite{zheng2023judging, li2024generation}, where an LLM automatically scores the synthetic interaction trajectories. 
This scored and ranked dataset serves as a foundation for reinforcement learning~\cite{ouyang2022training, christiano2017deep}, allowing the compact model to learn GUI interaction policies without requiring costly manual annotations. The pipeline of generating the data is shown in Figure~\ref{fig:cua} and illustrated in \S\ref{sec:method}.

We then fine-tune a 2B-model on our generated data using Direct Preference Optimization (DPO) training~\cite{wang2024mdpo} and evaluate our approach on the OS-World benchmark~\cite{xie2024osworld} with baseline. Our local agent outperforms the baseline significantly. These findings suggest that it is not only feasible but advantageous to shift toward local, privacy-aware computer use agents. Our work lays the groundwork for efficient, deployable, and generalizable GUI agents—capable of empowering a wide range of real-world applications.

\section{Related Work}
\label{sec:related_work}
\paragraph{Reinforcement Learning }
Reinforcement learning has become a central approach for aligning language and multimodal models with human preferences. 
\citet{christiano2017deep} introduce Reinforcement Learning from Human Feedback (RLHF), proposing a framework in which a reward model is trained using pairwise human preference comparisons.  This approach is later adapted to language generation tasks by \citet{ziegler2019fine}, demonstrating that preference-informed fine-tuning of large language models (LLMs) can significantly improve alignment with user expectations.
\citet{ouyang2022training} introduce a multi-stage pipeline: 
starting with supervised fine-tuning on demonstration data, followed by reward model training using human-labeled preference pairs, and finally optimizing the policy via Proximal Policy Optimization (PPO) \citep{schulman2017proximal}.
%supervised fine-tuning on demonstration data, reward model training from human-labeled preferences, and final optimization using Proximal Policy Optimization (PPO) \cite{schulman2017proximal}. 
The resulting InstructGPT models exhibits enhanced truthfulness and user satisfaction relative to larger, non-aligned baselines.
Despite PPO's effectiveness, its computational complexity and reliance on reward model rollouts have motivated the development of simplified alternatives. 
\citet{rafailov2023direct} propose DPO, which reframes preference-based alignment as a classification problem, thereby eliminating the need for explicit reward modeling or inference.
% Direct Preference Optimization (DPO), introduced by \citet{rafailov2023direct}, reframes preference optimization as a classification problem, avoiding explicit reward modeling and rollouts. 
Under DPO, the optimal policy is derived via a softmax over reward differences, enabling efficient end-to-end gradient-based training. Empirical results have demonstrated that DPO achieves comparable or superior alignment to PPO across tasks such as summarization, dialogue generation, and sentiment control, while offering greater training stability and reduced resource consumption.
Building on these insights, recent research has extended DPO into the multimodal domain \cite{li2023silkie, gunjal2024detecting}. 
\citet{zhang2024direct} apply a DPO-based strategy to video-language instruction tuning using GPT-4V-based preference signals. Addressing the challenge of unconditional preferences in multimodal learning, \citet{wang2024mdpo} propose mDPO, an extension that incorporates image-conditioned preference modeling and reward anchoring. Their approach reduces hallucinations and improved visual grounding across multiple vision-language models (VLMs).

\paragraph{Computer Use Agent Training}
The development of agents capable of interacting with graphical user interfaces (GUIs) has been significantly advanced by recent progress in multimodal learning. Such agents, often referred to as computer-use agents, are required to resolve key challenges such as long-horizon planning, multimodal grounding, partial observability, and efficient memory management in dynamic, temporally extended environments.
Early approaches in this area have targeted specific domains, including web navigation~\cite{deng2023mind2web, zhou2023webarena, zheng2024gpt}, desktop environments~\cite{bonatti2024windows, xie2024osworld, wu2024copilot}, and mobile platforms ~\cite{yan2023gpt, rawles2023androidinthewild}.
While effective within their respective domains, these systems often lacked generalization across diverse GUI distributions.
Recent methodological trends have shifted toward building general-purpose agents that abstract over specific interface modalities. For instance, AGUVIS~\cite{xu2024aguvis} and InfiGUIAgent~\cite{liu2025infiguiagent} propose two-stage training paradigms that explicitly decouple low-level GUI grounding from high-level task planning and reasoning. This modularization enables improved scalability and transferability across heterogeneous platforms.
UI-TARS~\cite{qin2025ui} introduces a self-reflective task-aware decision-making framework that employs DPO to refine action selection through recursive evaluation. Complementarily, Agent-S2~\cite{agashe2025agent} leverages a generalist-specialist architectural decomposition, in which a domain-agnostic planner is paired with a set of domain-specific execution modules. The inclusion of a Mixture-of-Grounding-Experts mechanism facilitates policy reuse across multiple interface distributions, contributing to improved sample efficiency and transfer learning.
Collectively, these approaches represent significant progress toward building general-purpose, robust multimodal agents. Despite these advances, state-of-the-art GUI agents predominantly rely on computationally intensive cloud-based inference services, which present substantial privacy and scalability limitations. Our work addresses these limitations by developing a lightweight vision-language model that can be operate entirely on local hardware, thereby preserving user privacy while maintaining robust performance under computational constraints.

\section{LLM-as-Judge DPO Training Pipeline}
\label{sec:method}
Improving the performance of computer use agents requires high-quality preference data, but collecting such data through human annotation is costly and difficult to scale. To address this, we propose an LLM-as-Judge framework that leverages GPT-4o \citep{openai2024gpt4o} to automatically assess and rank model-generated responses for UI-based tasks. Using this approach, we construct a scalable dataset of preference-labeled examples derived from the AGUVIS benchmark \citep{xu2024aguvis}, which consists of real-world GUI interaction tasks. These preference pairs are then used to fine-tune our policy model, UI-TARS-2B \citep{qin2025ui}, via Direct Preference Optimization (DPO) \citep{rafailov2023direct}, encouraging it to better emulate expert behavior in completing interface-based instructions. Figure~\ref{fig:cua} shows the overall pipeline of our data collection strategy and we will detail each step in the following. 

\paragraph{Data Sampling.}
To construct preference-labeled training data for DPO fine-tuning, we first sampled multiple responses from our base policy model, UI-TARS-2B, using task instances from the AGUVIS dataset. This dataset consists of natural language instructions paired with GUI screenshots, with the goal of predicting the next interaction step within the graphical interface. 
For each task instance, we generate a set of 10 diverse candidate completions using a hybrid decoding strategy that combines temperature-controlled nucleus sampling with top-\textit{k} filtering (temperature = 0.7, top \textit{k} = 10, top \textit{p} = 0.9).
This strategy constrains token selection to high-probability regions while introducing controlled randomness, striking a balance between output diversity and task relevance.

% For each task, we generated 10 diverse candidate completions using temperature-controlled nucleus sampling~\cite{holtzman2019curious} with top-k filtering (temperature=0.7, top\_k=10, top\_p=0.9). This hybrid decoding strategy balances diversity and relevance by restricting the candidate pool to high-probability tokens while allowing for some stochasticity. 

\paragraph{Response Scoring.}
To assess the quality of model-generated responses, we employ GPT-4o as an automatic evaluator. 
For each task in the AGUVIS dataset, we present GPT-4o with 10 sampled responses from UI-TARS-2B alongside the ground truth answer, all within a single prompt. 
We instruct GPT-4o to assign a quality score between 0 and 100 to each response based on its relevance, accuracy, and alignment with the task instruction. 
This setup enables GPT-4o to compare all responses in context and produce consistent, relative scores across candidates. Figure~\ref{fig:judge_prompt} shows the prompt to instruct the GPT-4o model. 
Note that the term \textit{instruction} refers to the input originally provided to UI-TARS-2B model during response sampling, which includes the task description, available action space, and the required output format.
%Note that the ``instruction'' is the input given to the UI-TARS-2B model when we sample the answer which includes the task, the action space, and the required output format.

\begin{figure}[htbp]
\centering
\begin{tcolorbox}[title=LLM-as-Judge Prompt Format, colback=gray!5, colframe=black!50, coltitle=black, fonttitle=\bfseries, sharp corners=southwest, width=0.45\textwidth]
You are a judger to grade multiple answers given an instruction and the ground truth answer. Please grade each candidate from 0 to 100.

\textbf{Instruction:} \texttt{\{instruction\}}

\textbf{Ground truth answer:} \texttt{\{gt\}}

\textbf{Candidate answers:} \texttt{\{ca\}}

Please output only a valid JSON object with two fields:
\begin{itemize}
    \item \texttt{"scores"}: A list of scores for each candidate.
    \item \texttt{"reason"}: A string explaining the reasoning behind the scores.
\end{itemize}
\end{tcolorbox}
\caption{Prompt used to obtain GPT-4o scores for candidate responses.}
\label{fig:judge_prompt}
\end{figure}

\paragraph{Pairing for DPO Training.}
We explore two strategies for constructing preference pairs based on the GPT-4o-generated scores.
The first strategy pairs each model-generated response with a score below 80 against the corresponding ground-truth answer.
In these pairs, the lower-scoring model output is treated as the \textit{rejected} sample, while the ground-truth answer serves as the \textit{preferred} one.
The second strategy constructs preference pairs directly between model-generated responses.
For each task, we compare model outputs with different scores and form pairs in which the higher-scoring response is designated as the \textit{preferred} sample and the lower-scoring one as the \textit{rejected} sample.
%In the second strategy, we directly compared model generated responses by forming pairs between answers with differing scores, treating the higher-scoring response as preferred and the lower-scoring one as rejected. 
These approaches enabled us to create a diverse set of preference pairs for DPO training.
Using the quality scores produced by GPT-4o, we construct these preference pairs and apply  DPO to further train the UI-TARS-2B model.
\section{Experiments}
\label{experiments}

\subsection{OSWorld Benchmark} 

The OSWorld benchmark is a diverse evaluation suite designed to assess the performance of computer use agents which comprises 369 real-world computer tasks involving web and desktop applications in open domains, OS file I/O, and workflows spanning  10 distinct domains (see Table~\ref{tab:results_15steps}.), including commonly used programs like Chrome, GIMP, LibreOffice tools (Calc, Impress, Writer), VS Code, VLC, and system-level operations (e.g., OS utilities and Thunderbird email). The  diversity is critical because it challenges agents to generalize beyond single-task performance and instead demonstrate robustness, flexibility, and adaptability across heterogeneous user interfaces and workflows.

\subsection{Experiments Setup}
\paragraph{Model Inference Parameters Setup} For all models, we set the temperature parameter to 1.0, and top\_p to 0.7, and the maximum number of tokens for generation is set to 1500 (while in practice, models generate much shorter length than this maximum value). Given the variance of the generated output due to the temperature, for each models, we run the evaluation three times and average the performance. 
\paragraph{Hardware Setup}
To comprehensively evaluate and benchmark multiple  vision-language models in parallel, we leverage a scalable infrastructure hosted on the Intel\textsuperscript{\tiny{\circledR}} Tiber AI Cloud. Our environment consists of five dedicated servers, each with dual Intel\textsuperscript{\tiny{\circledR}} Xeon\textsuperscript{\tiny{\circledR}} Platinum 8468 Processors, 1TB of RAM, connected to a high-performance network file system, as well as several local hosted vLLM inference endpoints. This robust infrastructure allows us to execute over 140 concurrent evaluations simultaneously. 
By optimizing the parallel execution of jobs and employing multiple local vLLM inference endpoints, we achieve efficient throughput and scalability. This setup enables the completion of all 369 examples in under 30 minutes for 15-step interactions and 90 minutes for 50-steps, significantly reducing the total evaluation time compared to traditional single node setups. 

% We utilize Kubernetes as our orchestration platform, organizing individual evaluations into distinct pods. Each pod comprises two containers: one hosting the OSWorld Ubuntu virtual machine, while the other runs the agent interfacing with the VM. Jobs in Kubernetes orchestrate multiple pods concurrently, enabling parallel evaluations across the entire OSWorld benchmark suite of 369 examples. 

\subsection{GPT-4 Scoring Baseline Analysis}

\begin{figure}[htbp]
\centering
\includegraphics[width=0.45\textwidth]{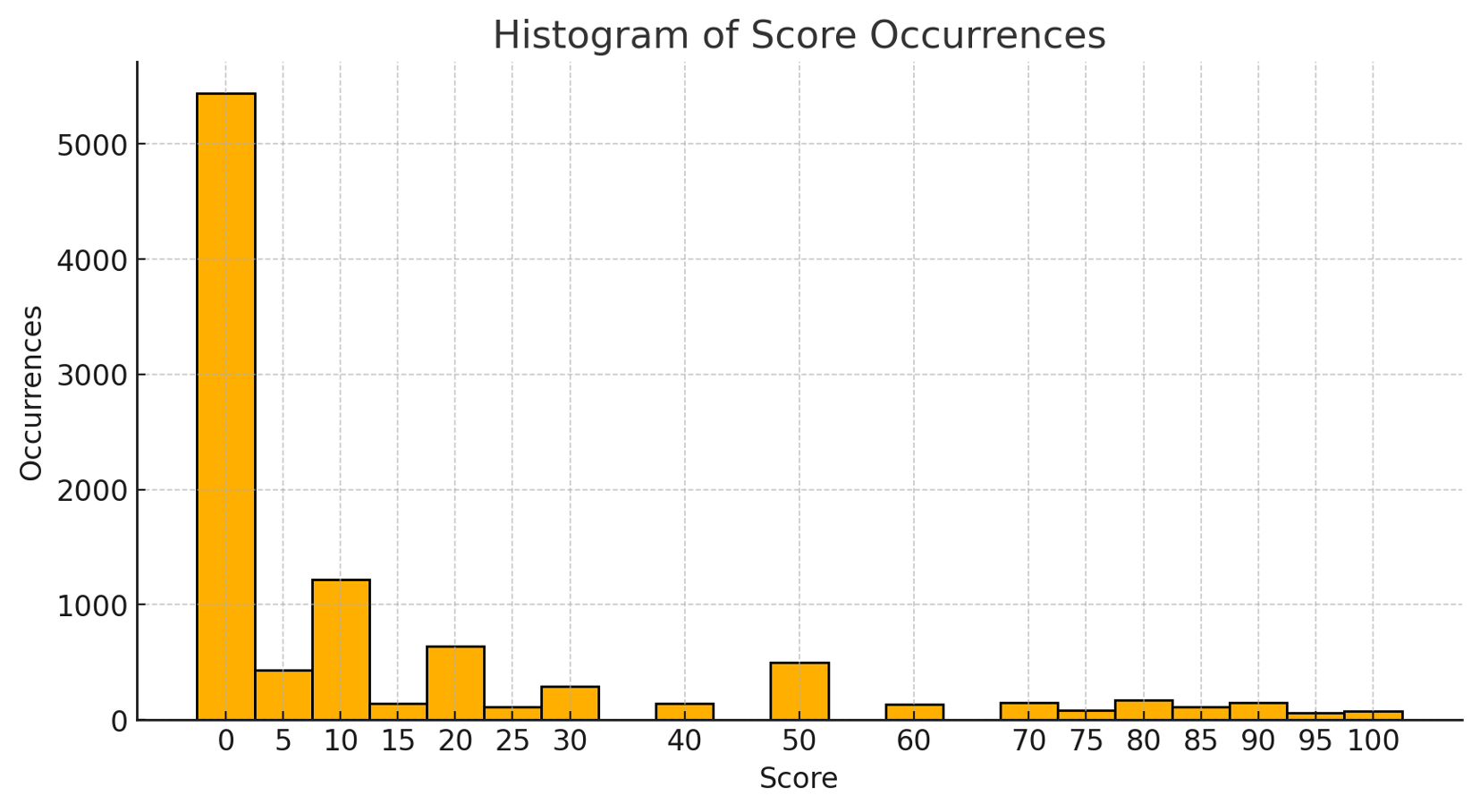}
\caption{LLM-as-judge Score distribution.}
\label{fig:llm-scores}
\end{figure}

Figure~\ref{fig:llm-scores} shows the distribution of GPT-4o scores on responses generated by the baseline UI-TAR-2B model. The scores are heavily concentrated near zero, with the vast majority of samples receiving a score of 0 and only a small fraction scoring between 5 and 50. Responses above 50 are extremely rare. This distribution highlights that the baseline model, despite being state-of-the-art for its size, frequently generates low-quality outputs and still has significant room for improvement. 
The abundance of poor responses enables effective preference pair construction when contrasted with the ground truth, forming the basis for learning through DPO.

\subsection{Benchmark Experiments Results} 

We refer to the models trained using the first and second pairing strategies as DPO-1 and DPO-2, respectively, and the model trained on their combined dataset as DPO-3. We conduct experiments under two settings: \textbf{15-Steps}, where the agent can take up to 15 actions to complete a task, and \textbf{50-Steps}, where the limit is increased to 50. We analyze the results for each setting separately.

\paragraph{15-Steps Results}
Table~\ref{tab:results_15steps} compares our three DPO models with the baseline under the 15-step setting. All DPO models outperform the baseline, with weighted average improvements of 32.86\%, 22.43\%, and 35.03\% for DPO-1, DPO-2, and DPO-3, respectively. Notably, different DPO models excel in different domains—DPO-3 leads in Chrome and Thunderbird, while DPO-2 outperforms others in OS and VLC—indicating complementary strengths across strategies.  DPO-3 achieves the highest overall score (9.33) and delivers the best performance in 4 out of 10 domains. Overall, DPO-3 demonstrates the most consistent and robust gains, validating the benefit of integrating both preference supervision schemes.

\paragraph{50-Steps Results}
Table~\ref{tab:results_50steps} presents the model performance under the 50-step setting. Although DPO-1 leads in 4 out of 10 domains, the overall performance gains across all DPO models are relatively modest in this setting. Interestingly, we observe that all DPO models perform worse under the 50-step setting compared to their 15-step counterparts. One possible explanation is that longer action horizons increase the likelihood of error accumulation, making it harder for the model to remain aligned with task goals—especially if training was more biased toward shorter trajectories. Additionally, our system did experience hardware failures during inference in some 50-step runs, which could have contributed to degraded performance. We plan to investigate both hypotheses further by analyzing failure cases and rerunning affected evaluations under controlled conditions. 
Nevertheless, a noteworthy finding is that DPO models in the 15-step setting still outperform the baseline even when it operates under the more flexible 50-step setting. This demonstrates that DPO fine-tuning can lead to more efficient and reliable behavior with shorter interaction budgets, providing practical benefits in both performance and inference efficiency.
\begin{table}[ht]
\centering
\caption{Success rates of our models vs. baseline across 10 application domains in OS-World benchmark. Each agent has maximum \textbf{15} steps. LO stands for LibreOffice. Success rates are calculated independently for each domain. Max Diff is computed based on the best performance compared to the average baseline}
\label{tab:results_15steps}
\footnotesize
\resizebox{\linewidth}{!}{%

\begin{tabular}{@{}lccccc@{}}
\toprule
\textbf{Domain} & \textbf{DPO-1} &\textbf{DPO-2} & \textbf{DPO-3} & \textbf{Baseline} & \textbf{Max Diff.} \\
\midrule
Chrome & 10.87 & 7.97 & \textbf{12.43} & 5.07 & 10.48 \\
GIMP & 29.49 & 24.36 & 27.28 & \textbf{30.77} & 7.69 \\
LO Calc & \textbf{3.55} & 2.13 & \textbf{3.55} & 2.13 & 4.26 \\
LO Impress & \textbf{4.96} & 4.27 & 3.59 & 3.59 & 2.93 \\
LO Writer & 5.79 & \textbf{7.24} & 5.80 & 4.35 & 4.35 \\
Multi Apps & 0.66 & 1.32 & 1.17 & \textbf{1.32} & 0.66 \\
OS & 22.22 & \textbf{27.78} & \textbf{27.78} & 19.44 & 13.89 \\
Thunderbird & 8.89 & \textbf{13.33} & \textbf{13.33} & 6.67 & 13.33 \\
VLC & 16.50 & \textbf{10.40} & 8.21 & 1.96 & 21.57 \\
VS Code & \textbf{24.24} & 19.70 & 23.78 & 18.18 & 12.25 \\
\midrule 
Weighted Avg. & 9.18 & 8.46 & 9.33 & 6.91 & 3.14 \\
\# Best of Domain & 3 & 4 &4 &2 & - \\
\% improvement & 32.86 & 22.43 & 35.03 & - & - \\
\bottomrule
\end{tabular}
}
\end{table}

\begin{table}[ht]
\centering
\caption{Success rates of our models vs. baseline across 10 application domains in OS-World benchmark. Each agent has maximum \textbf{50} steps. LO stands for LibreOffice. Success rates are calculated independently for each domain.}
\label{tab:results_50steps}
\resizebox{\linewidth}{!}{%
\begin{tabular}{@{}lccccc@{}}
\toprule
\textbf{Domain} & \textbf{DPO-1} & \textbf{DPO-2} & \textbf{DPO-3} & \textbf{Baseline} & \textbf{Max Diff.} \\
\midrule
Chrome & \textbf{9.37} & 7.25 & 8.64 & 5.07 & 9.99 \\
GIMP & 28.62 & 30.77 & 24.36 & \textbf{31.18} & 3.44 \\
LO Calc & 2.13 & \textbf{4.26} & 2.85 & 2.84 & 1.42 \\
LO Impress & 2.90 & 3.58 & \textbf{5.01} & 4.26 & 2.19 \\
LO Writer & \textbf{11.59} & 7.24 & 8.82 & 4.34 & 8.70 \\
Multi Apps & \textbf{1.32} & 0.99 & 1.17 & 0.99 & 0.99 \\
OS & \textbf{27.78} & 25.00 & 20.83 & 26.39 & 6.94 \\
Thunderbird & 11.11 & 6.67 & 15.56 & \textbf{20.00} & 0.00 \\
VLC & 7.97 & 14.22 & 6.70 & \textbf{15.27} & 2.37 \\
VS Code & 24.00 & \textbf{25.03} & 19.37 & 21.94 & 8.50 \\
\midrule 
Weighted Avg & 9.01 & 8.88 & 8.24 & 8.83 & 1.52 \\
\# Best of Domain & 4 & 2 & 1 & 3 & - \\
\% improvement & +2.04\% & +0.57\% & -6.68\% & - & - \\
\bottomrule
\end{tabular}%
}
\end{table}

\subsection{Failure Case Analysis}
To better understand the limitations of models, we conduct a qualitative analysis of failure cases by manually inspecting samples where the agent failed to complete the task. We categorize the observed failures into three main groups:

\paragraph{Invalid Coordination.}
In this category, the model generates syntactically incorrect actions that cannot be executed by the parsing system (e.g. PyAutoGUI). For example, the model may output a coordinate in an incorrect format such as ''click(start\_box='[19,60)')''. which yield parsing error ``could not convert string to float: '[19'''. Although the intended target may be correct, the formatting error renders the action invalid.

\paragraph{Wrong Coordination.}
In this case, the model generates syntactically valid but semantically incorrect coordinates. The predicted location does not correspond to the intended UI element, leading the agent to click on irrelevant or empty areas. These errors suggest gaps in spatial grounding of visual context in the screenshot.

\paragraph{Repeat Wrong Action.}
In this failure mode, the model produces an invalid or incorrect action and then gets stuck repeating the same faulty step across multiple timesteps. This behavior indicates insufficient corrective mechanisms during generation, where the model fails to revise its plan even after repeated failure signals.

These categories highlight key areas for improvement in model robustness, including syntax enforcement, spatial grounding, and recovery from invalid states. We plan to explore explicit action validation, visual feedback mechanisms, or trajectory-level supervision to mitigate these issues in the future.
\section{Conclusion}
We propose an LLM-as-Judge framework to improve lightweight, locally-executed computer use agents without requiring human-labeled data. By leveraging GPT-4 to automatically evaluate model-generated interaction trajectories, we construct high-quality preference pairs for Direct Preference Optimization. Our experiments on the OS-World benchmark show that DPO-trained models consistently outperform the baseline, particularly in short action trajectory (maximum 15 steps) settings where efficient interaction is critical. Moreover, our qualitative analysis reveals actionable failure patterns, offering guidance for future improvements in coordination accuracy and recovery behavior. This work demonstrates a scalable and privacy-preserving approach to training compact GUI agents, paving the way toward more capable and generalizable systems deployable on personal devices.
{
    \small
    \bibliographystyle{ieeenat_fullname}
    \bibliography{main}
}


\end{document}